\documentclass[letterpaper, 10 pt, conference]{ieeeconf} 
\usepackage{url}
\usepackage{graphics}
\usepackage{graphicx}
\usepackage{caption}
\usepackage{subcaption}
\usepackage{algorithmic}
\usepackage{algorithm}
\usepackage{setspace}
\usepackage{epsfig}
\usepackage{amsfonts}
\usepackage{multirow}
\usepackage{todonotes}
\usepackage{pbox}

\usepackage{array}

\usepackage{amssymb,url,amsmath}
\usepackage{algorithm, algorithmic}
\usepackage{graphicx}
\usepackage{color}
\usepackage{multirow}
\usepackage{balance}

\usepackage{booktabs}       
\usepackage{amsfonts}       
\usepackage{nicefrac}       
\usepackage{microtype}      
\usepackage{caption}
\usepackage{booktabs}
\usepackage{tikz}
\usepackage{array}
\usepackage{nicefrac}

\usepackage{color}
\definecolor{CommentRed}{rgb}{0.7,0,0}
\definecolor{CommentBlue}{rgb}{0,0,0.7}
\definecolor{CommentGreen}{rgb}{0,0.7,0}

\newcolumntype{L}[1]{>{\raggedright\let\newline\\\arraybackslash\hspace{0pt}}m{#1}}
\newcolumntype{C}[1]{>{\centering\let\newline\\\arraybackslash\hspace{0pt}}m{#1}}
\newcolumntype{R}[1]{>{\raggedleft\let\newline\\\arraybackslash\hspace{0pt}}m{#1}}

\newcommand{\mypm}{\mathbin{\smash{%
\raisebox{0.35ex}{%
            $\underset{\raisebox{0.5ex}{$\smash -$}}{\smash+}$%
            }%
        }%
    }%
}

\definecolor{myblue}{RGB}{0,51,102}
\IEEEoverridecommandlockouts                              
\title{\LARGE \bf Addressing Appearance Change in Outdoor Robotics \\ with Adversarial Domain Adaptation  }

\author{Markus Wulfmeier$^{1}$, Alex Bewley$^{1}$ and Ingmar Posner$^{1}$
\thanks{$^{1}$The authors are with the Oxford Robotics Institute, Department of Engineering Science, 
        University of Oxford, United Kingdom; \newline 
        {\tt\small markus, bewley, ingmar@robots.ox.ac.uk}}}%

\begin{document}

\maketitle
\thispagestyle{empty}
\pagestyle{empty}

\maketitle

\begin{abstract}
Appearance changes due to weather and seasonal conditions represent a strong impediment to the robust implementation of machine learning systems in outdoor robotics. While supervised learning optimises a model for the training domain, it will deliver degraded performance in application domains that underlie distributional shifts caused by these changes. Traditionally, this problem has been addressed via the collection of labelled data in multiple domains or by imposing priors on the type of shift between both domains. We frame the problem in the context of unsupervised domain adaptation and develop a framework for applying adversarial techniques to adapt popular, state-of-the-art network architectures with the additional objective to align features across domains. Moreover, as adversarial training is notoriously unstable, we first perform an extensive ablation study, adapting many techniques known to stabilise generative adversarial networks, and evaluate on a surrogate classification task with the same appearance change. The distilled insights are applied to the problem of free-space segmentation for motion planning in autonomous driving.

\end{abstract}

\section{Introduction}
\label{sec:introduction}

In this paper we revisit the prominent issue in field robotics of appearance change under the influence of numerous factors including time of day, weather, and seasonal variation. Dealing with these changes becomes relevant in all modules of the robotic perception system including localisation~\cite{paton2015s}, mapping~\cite{ChurchillIJRR2013} and obstacle detection~\cite{maddern2014illumination}.
Given the recent adoption of high capacity deep neural networks for many robotic vision tasks, the effects of condition variation have been moderately alleviated with the assumption that the labelled training data is diverse enough to capture the variation expected during deployment. However, due to the expense and impracticality of collecting labels across all environmental conditions, the training data is commonly captured in a subset of episodes which exacerbates the effects of appearance change when deployed.

\begin{figure}[t]
	\centering
		\includegraphics[width = 0.47\textwidth]{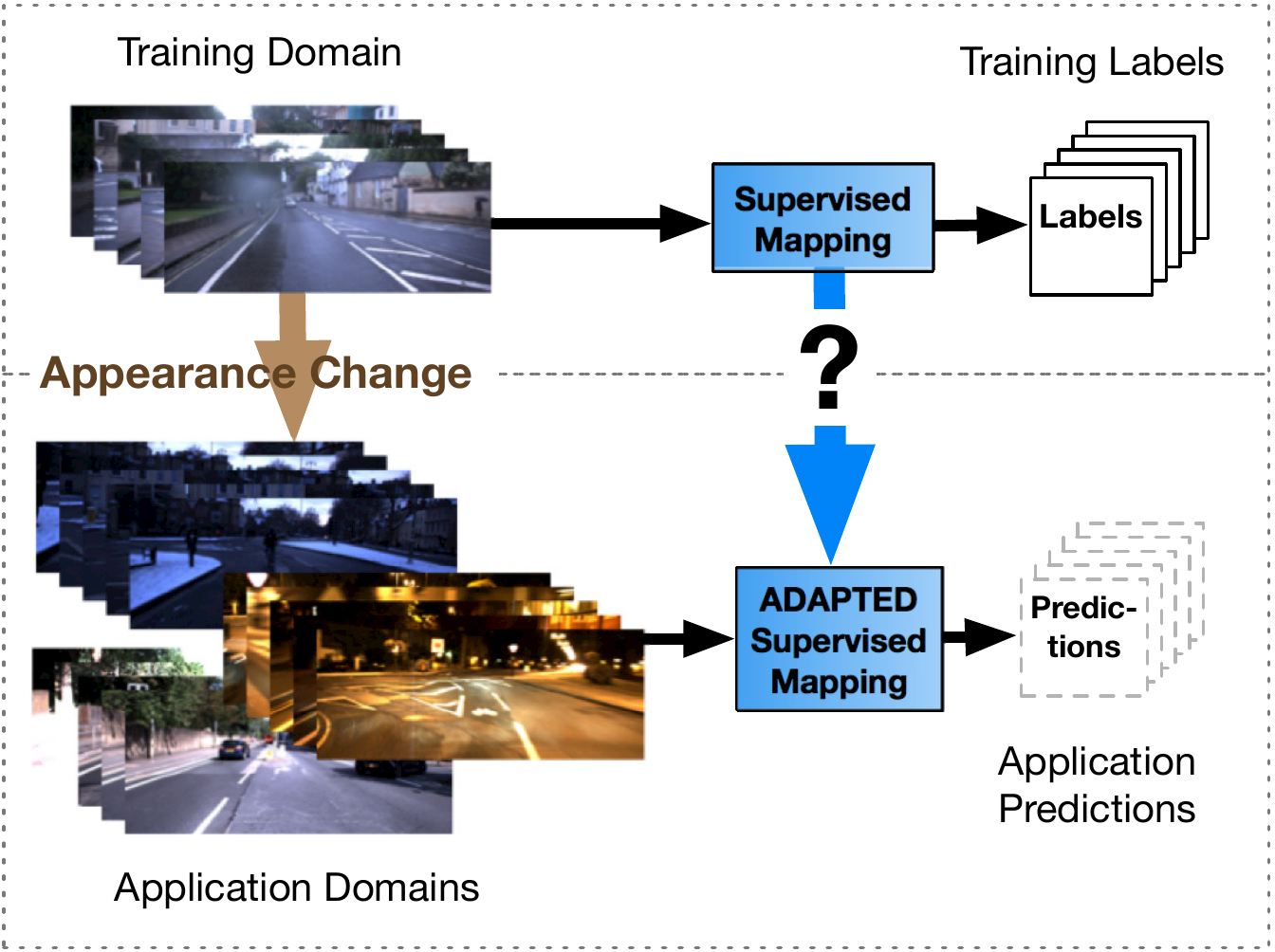}
    \caption{\small Appearance change between training and test time of machine learning systems. Adversarial domain adaptation (ADA) provides a robust, systematic way to adapt the supervised task to perform in the application domains without supervised label information.}
	\label{fig:appearance_change}
\end{figure}

This work addresses the problem of appearance change by reframing it in the context of unsupervised domain adaptation. 
With this view, we are acknowledging that the underlying distribution of our labelled training data may differ considerably from unlabelled data encountered in the application domain.
Methods for domain adaptation have already found success in robotics for transfer from 3D models to laser data \cite{lai20093d} or from simulation to real images collected indoors \cite{tzeng2015towards}. Here we focus on developing a general and flexible framework for adapting supervised machine learning models to address appearance change for outdoor robotics problems (as depicted in Figure \ref{fig:appearance_change}).

Traditional approaches for addressing the general domain adaptation problem have focused on modelling the density of the source and target distributions separately~\cite{daume2006domain} or with imposed prior structure~\cite{sugiyama2008direct}. However, density estimation itself is a challenging issue, particularly for appearance, as it is difficult to impose a prior without making assumptions on the distribution. Recently, generative adversarial networks (GAN) were proposed~\cite{goodfellow2014generative} as a framework to model any arbitrary distribution where a generating network is optimised to produce data indistinguishable from real data as considered by a discriminating network. Building on the flexibility of GANs, adversarial domain adaptation (ADA) has demonstrated astonishing performance for unsupervised domain adaption~\cite{ShrivastavaPTSW16, bousmalis2016domain}.

Extending existing work~\cite{Ganin2016, bousmalis2016domain}, we demonstrate a straightforward framework for the adaptation of existing, commonly used network architectures that allows to benefit from recent progress in deep learning and crystallise the principal factors of influence on target domain performance. 
To address real world applicability of ADA with this schema, we modify two popular network architectures, AlexNet~\cite{krizhevsky2012imagenet} and FCN-VGG16~\cite{long2015fully} for the tasks of classification and pixel-wise image segmentation respectively.  
When considered in an outdoor context, both tasks are affected by change in appearance caused by factors such as time of day or weather conditions.

Due to the documented evidence of training instability of adversarial training procedures \cite{goodfellow2014generative, Goodfellow17, arjovsky2017towards}, 
we first use place classification as a surrogate task to perform an extensive evaluation of approaches for balancing our objectives and stabilising the training process. 
The reduction in complexity enables us to enumerate and explore various potential configurations of the network structure and training procedures to identify the key factors of influence before scaling up to the more challenging segmentation task. Particularly, we seek to answer the following questions:
\begin{enumerate}
    \item Trade-offs: How do we stabilise training and balance supervised and adversarial objectives?
    \item Architecture: How can we directly adapt existing network architectures that are known to perform well?
    \item Performance: Where does ADA work best and how do larger appearance changes influence performance?
\end{enumerate}
A series of experiments is presented to provide valuable insights around the above questions facilitating the extension to full image pixel-wise traversability segmentation commonly used as an input to motion planning systems. In the final segmentation task we see a significant performance increase in the target domain by applying ADA over the FCN-VGG16~\cite{long2015fully} baseline. 
We benefit from applying the confusion loss method for training generative adversarial networks \cite{goodfellow2014generative} and furthermore investigate a patch-wise discriminator schema for the segmentation task.
Finally, we also demonstrate the method's pertinence as a regulariser by observing performance gains within the original domain as more unlabelled data becomes available from similar domains.

\vspace{2mm}

\section{Related Work}
\label{sec:related}

In robotics, the problem of appearance change has long been an issue for both laser and image data.
Lai and Fox \cite{lai20093d} presented an exemplar based approach to align spin-image features from web based 3D models to dense laser scan.
For visual appearance, the colour-constant transform \cite{ratnasingam2010study} is commonly applied directly on the image to achieve lighting invariance for localisation \cite{corke2013dealing}, visual teach and repeat \cite{paton2015s} and segmentation \cite{maddern2014illumination}. Other works in localisation take a multi-view approach by either exploiting temporal structure \cite{milford2012seqslam} or by accumulating multiple experiences covering different appearances \cite{ChurchillIJRR2013} to avoid the issue of modelling appearance change. The work of Neubert \emph{et al.} \cite{neubert2013appearance} is the closest to our approach where they learn to synthesise stored images into the current season. By casting the appearance change issue as an unsupervised domain adaption problem our approach differs from the previous work as we do not require known cross domain correspondences.

In the context of domain shift between training and test data, Ben-David \emph{et al.} \cite{ben2007analysis} derived theoretical upper bounds on a classifiers performance.
While unsupervised domain adaptation is an open research problem in theoretical and practical terms \cite{bousmalis2016domain, bousmalis2016unsupervised}, recent successes have shown the capability to train expressive, flexible models and address high dimensional input distributions and therefore made first steps in enabling real world applicability \cite{ShrivastavaPTSW16,Hong2017}.

The majority of recent successes have built on the empirical superiority of Neural Networks that enable hierarchical representation learning with respect to arbitrary combinations of differential objectives. In the context of domain adaptation we can minimise the mutual information between feature representations and the underlying domain while maximising relevance towards a supervised objective in the source domain.
Long \emph{et al}~\cite{Long0J16a} focus on minimising the Maximum Mean Discrepancy for the feature distributions of multiple layers of the network architecture. Sun \emph{et al.} \cite{SunFS16} align second order statistics of layer activations for source and target domain.

Recently, the field has been extended with adversarial methods to domain adaptation~\cite{Ganin2016, ajakan2014domain} which have resulted in strong performance benefits \cite{bousmalis2016domain}.
Adversarial unsupervised domain adaptation relies on training a discriminator to differentiate the domains underlying feature distributions while competing against an encoder which attempts to conceal the origin domain of input samples. 
While the loss formulation for the discriminator stays consistent between most approaches, different objectives have been applied for the encoder, including the \textit{minimax} formulation resulting from a `gradient reversal layer' \cite{Ganin2016}, and the \textit{confusion loss} \cite{tzeng2015simultaneous} that was found to address problems with vanishing gradients based on discriminator saturation in Generative Adversarial Networks (GAN) \cite{goodfellow2014generative}.
These methods have recently been employed to address tasks in medical image segmentation \cite{KamnitsasBLNSKM16}, gaze estimation \cite{ShrivastavaPTSW16} and transfer for reinforcement learning \cite{tzeng2015towards}.

\vspace{2mm}
\section{Adversarial Domain Adaptation}
\label{sec:ada}

The principal goal of ADA is to maximise the performance of a supervised task not only in the source domain - where labels are available - but also in the unlabelled target domain. As a domain shift potentially exists between training and application domains, the approach tries to align the marginal feature distributions independent of label for both domains to achieve its goal.

Therefore, we train the supervised task module and encoder to maximise the likelihood of source labels given the source inputs. 
Additionally, to adapt the model towards performing well in the target domain, we train the encoder to confuse a discriminator which tries to estimate the domain of a data sample and serves as density model. 

\begin{figure}[t]
	\centering
	\begin{tabular}{c c}		
	    \includegraphics[width = 0.25\textwidth]{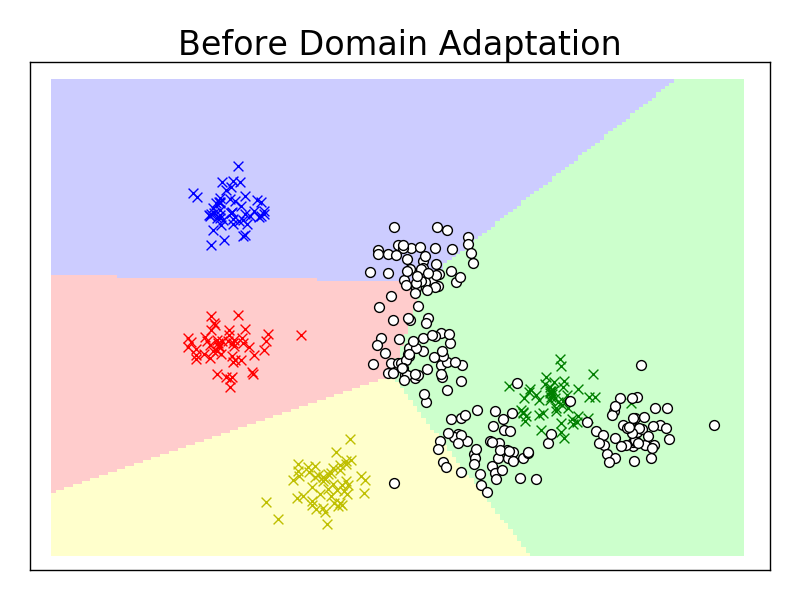}&\hspace{-5mm}
        \includegraphics[width = 0.25\textwidth]{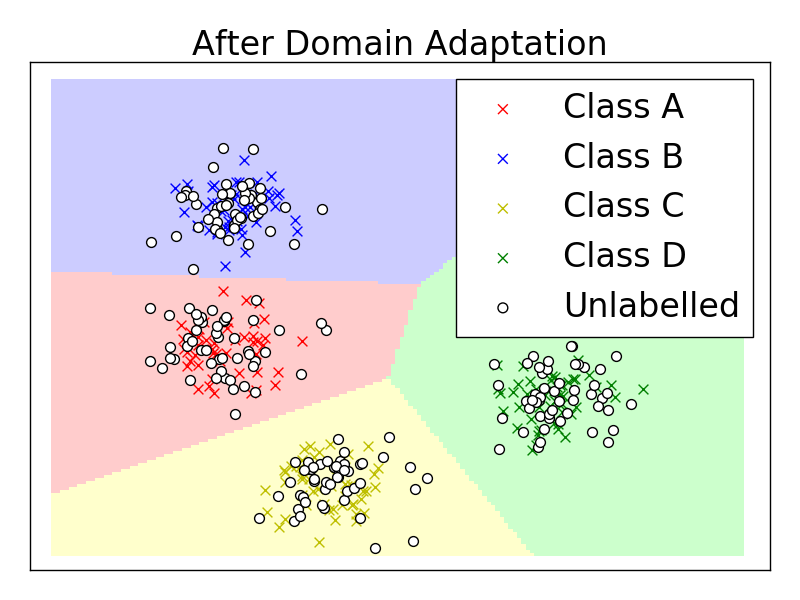}\\
	\end{tabular}
    \caption{\small An illustration of domain shift where the distribution of the unlabelled target domain is misaligned with the labelled source domain but both distributions share structural similarities. By aligning the two domains the decision boundary of the supervised task (denoted in colour) better aligns across both domains.}
	\label{fig:distribution}
\end{figure}

The adversarial training process works towards aligning the marginal feature distributions of source and target domains. With both domains sharing partial structural similarity, this also implicitly aligns the conditional distributions given the labels. Consequently, by aligning the distributions, ADA increases the performance of a supervised module with decision boundaries optimised for the source domains as displayed in Figure \ref{fig:distribution}.

\begin{figure}[t]
	\centering
		\includegraphics[width = 0.5\textwidth]{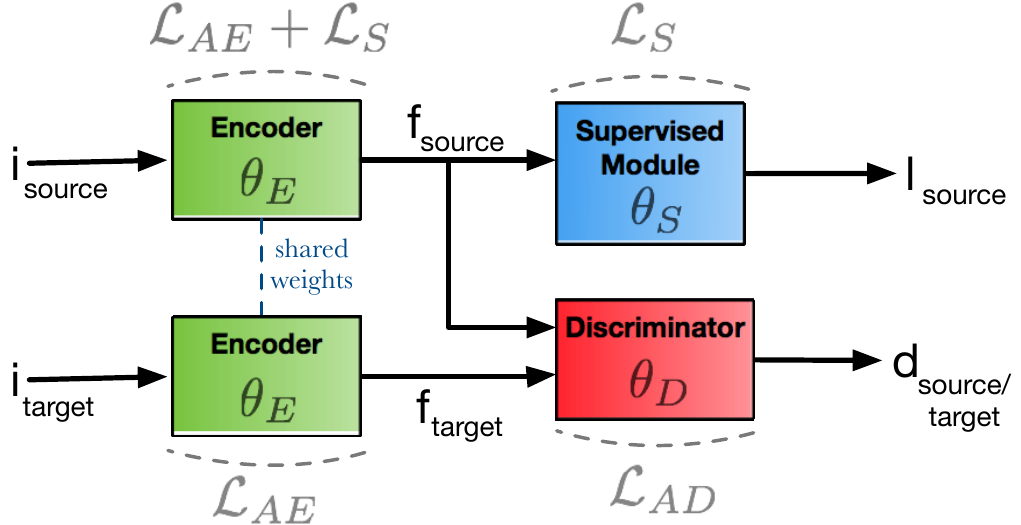}
    \caption{\small Schema for adversarial domain adaptation. Given a well-established network architecture, we determine a split layer and pass the activations of that layer additionally to a discriminator module. While the supervised module is only applied to source data, both domains share the same encoder.}
	\label{fig:ada}
\end{figure}

An existing network architecture is divided into two functional blocks referred to as \textit{encoder} and \textit{supervised task} as represented in Figure \ref{fig:ada}. Let $E: \mathbb{R}^n \rightarrow  \mathbb{R}^m $ be the encoder that transforms an input image $i_i$ into the feature representation $f_i$, which subsequently serves as input for both the supervised task $S: \mathbb{R}^m \rightarrow  \mathbb{R}^c $ producing label $l_i$ and the adversarial domain discriminator $D: \mathbb{R}^m \rightarrow  \mathbb{R} $ computing domain label $d_i$. 
To reduce the memory footprint of the model, we can remove the discriminator at test time as it is only an auxiliary module to determine the encoder objective.

The model is optimised to simultaneously minimise supervised and adversarial losses, respectively $\mathcal{L}_S$ and  $\mathcal{L}_A$. The adversarial loss is furthermore divided into the loss applying to the encoder $\mathcal{L}_{AE}$ and discriminator $\mathcal{L}_{AD}$ (see Equations \ref{eq:ada1} and \ref{eq:ada3}). The supervised objective is minimising the cross-entropy loss from Equation \ref{eq:ada2}.
While the adversarial encoder and discriminator losses both depend on encoder parameters $\theta_E$ and discriminator parameters $\theta_D$, each loss is only applied to the corresponding module to realise the adversarial training procedure. The filters of the supervised module $\theta_S$ are only optimised with respect to the supervised loss on data only from the source domain. The factor $\lambda$ determines the relative strength of supervised and adversarial objective.
In this arrangement, the encoder is encouraged to extract features that balance the relevance of the supervised task and the maximisation of domain invariance.

\begin{eqnarray}\label{eq:maxent_objective}
    \mathcal{L}(\boldsymbol \theta_S,\boldsymbol \theta_D,\boldsymbol \theta_E) &=& \mathcal{L}_S(\boldsymbol \theta_S,\boldsymbol \theta_E) + \lambda \mathcal{L}_{A}(\boldsymbol \theta_D, \boldsymbol \theta_E) \label{eq:ada1}\\
    \mathcal{L}_S(\boldsymbol \theta_S,\boldsymbol \theta_E) &=& \mathbb{E}_{l=S(E(i,\theta_E),\theta_S), i\sim S }[- \log(l)]\label{eq:ada2} \\ 
    \mathcal{L}_A(\boldsymbol \theta_D, \boldsymbol \theta_E) &=&  \mathcal{L}_{AD}(\boldsymbol \theta_D)  +  \mathcal{L}_{AE}(\boldsymbol \theta_E)  \label{eq:ada3} \\
    \mathcal{L}_{AD}(\boldsymbol \theta_D) &=& \mathbb{E}_{f=E(i), i\sim S }[- \log(D(f,\theta_D))]  + \label{eq:discr_loss}\\
    \nonumber &&\mathbb{E}_{f=E(i), i\sim T}[- \log(1 - D(f,\theta_D))]
\end{eqnarray}

The adversarial encoder loss $\mathcal{L}_{AE}$ is characterised independently of the discriminator loss with two common formulations considered in this work.
As with the original generator loss in the GAN framework  \cite{goodfellow2014generative}, the encoder in ADA can be trained to maximise the discriminator’s domain confusion \cite{tzeng2015simultaneous}:

\begin{eqnarray}
    \nonumber \text{\textcolor{gray}{confusion loss}} &&\text{\textcolor{gray}{:}}\\
    \mathcal{L}_{AE}(\boldsymbol \theta_E) &=& - \mathbb{E}_{f=E(i,\theta_E), i\sim S }[\log(1-D(f))]  \label{eq:conf_gen_loss}\\
    \nonumber && - \mathbb{E}_{f=E(i,\theta_E), i\sim T}[\log(D(f))]  . 
\end{eqnarray}

Contrary to the GAN framework, this loss applies to samples from both domains instead of only applying to generated samples.
An alternative is the minimax formulation which simply negates the discriminator loss~\cite{bousmalis2016unsupervised,ShrivastavaPTSW16}. 
When applied with gradient reversal \cite{Ganin2016}, the minimax loss can be expressed as:

\begin{eqnarray}
    \nonumber \text{\textcolor{gray}{minimax loss}} &&\text{\textcolor{gray}{:}}\\
    \mathcal{L}_{AE}(\boldsymbol \theta_E) &=& - \mathcal{L}_{AD} \label{eq:orig_gen_loss} \\
    \nonumber &=& \mathbb{E}_{f=E(i,\theta_E), i\sim S }[\log(D(f))]  \\
    \nonumber && + \mathbb{E}_{f=E(i,\theta_E), i\sim T}[\log(1 - D(f))] . 
\end{eqnarray}

Both the confusion and minimax losses are evaluated in Section \ref{sec:tradeoffs}.

\vspace{1mm}
\section{Ablation Study}
\label{sec:tricks}

Unsupervised domain adaptation is an inherently complex task as the lack of labelled information complicates the alignment between feature representations from source and target domain. 
While a high capacity encoder such as a deep neural network is per se capable of modelling even complex relations, the process of alignment might remove important information if it simplifies the alignment.
The difficulty of the process is proportional to the difference between the underlying source and target domains.
The field of autonomous driving presents a strong opportunity here as the use of overlapping routes can provide a partial alignment which simplifies the overall learning process.

We focus for our main evaluation on the surrogate task of classification and extend the evaluation subsequently to image segmentation with focus on path proposals for autonomous driving \cite{BarnesMP16}. The hyperparameter study is built on a small subset of labelled data from the publicly available \emph{Oxford RobotCar Dataset} consisting of over 1000 km of driving data with corresponding images, LIDAR, GPS and INS data \cite{RobotCarDatasetIJRR}.

\begin{figure}[t]
	\centering
	\begin{tabular}{c}
		\includegraphics[width = 0.43\textwidth]{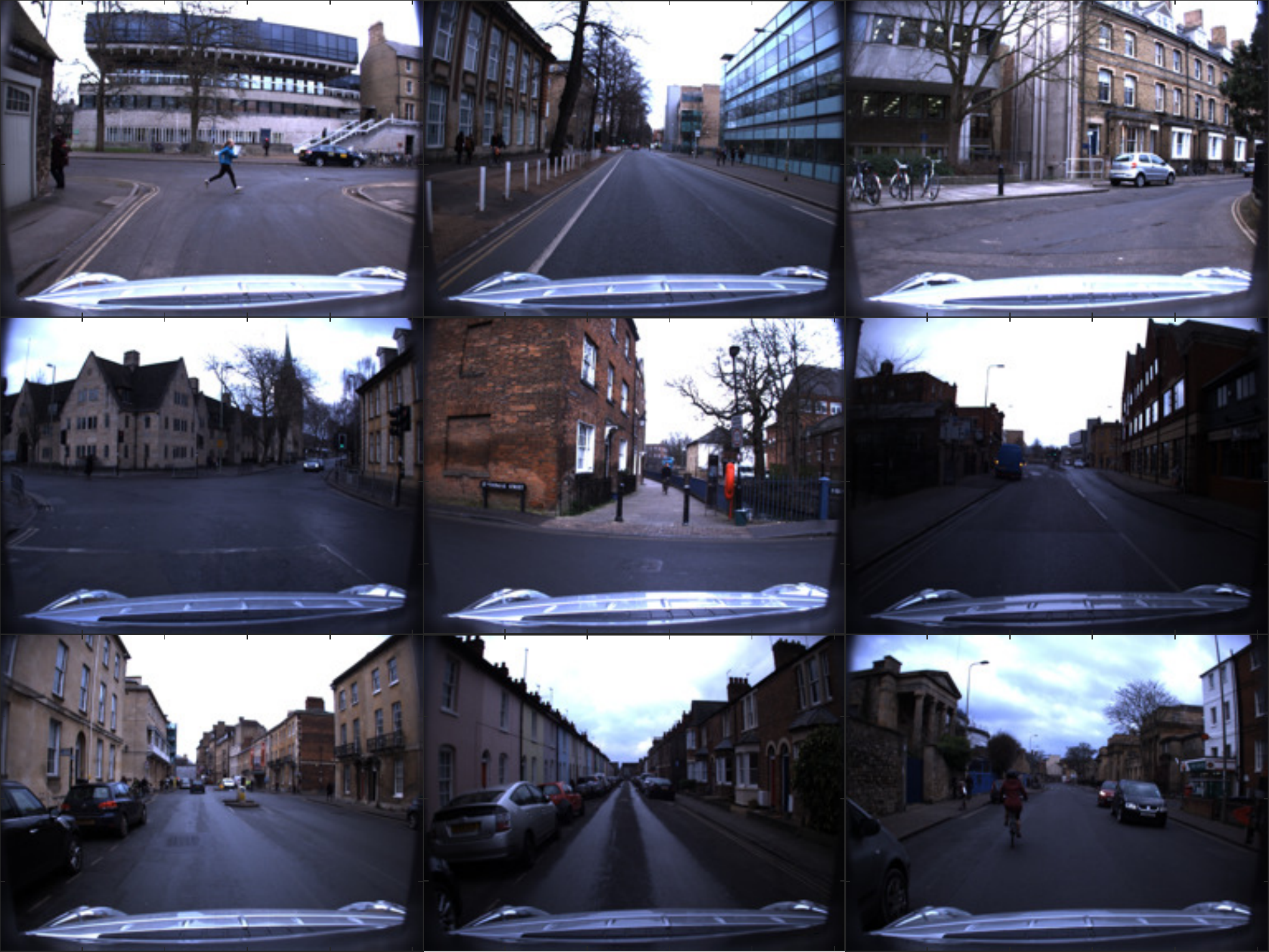}\\
        \includegraphics[width = 0.43\textwidth]{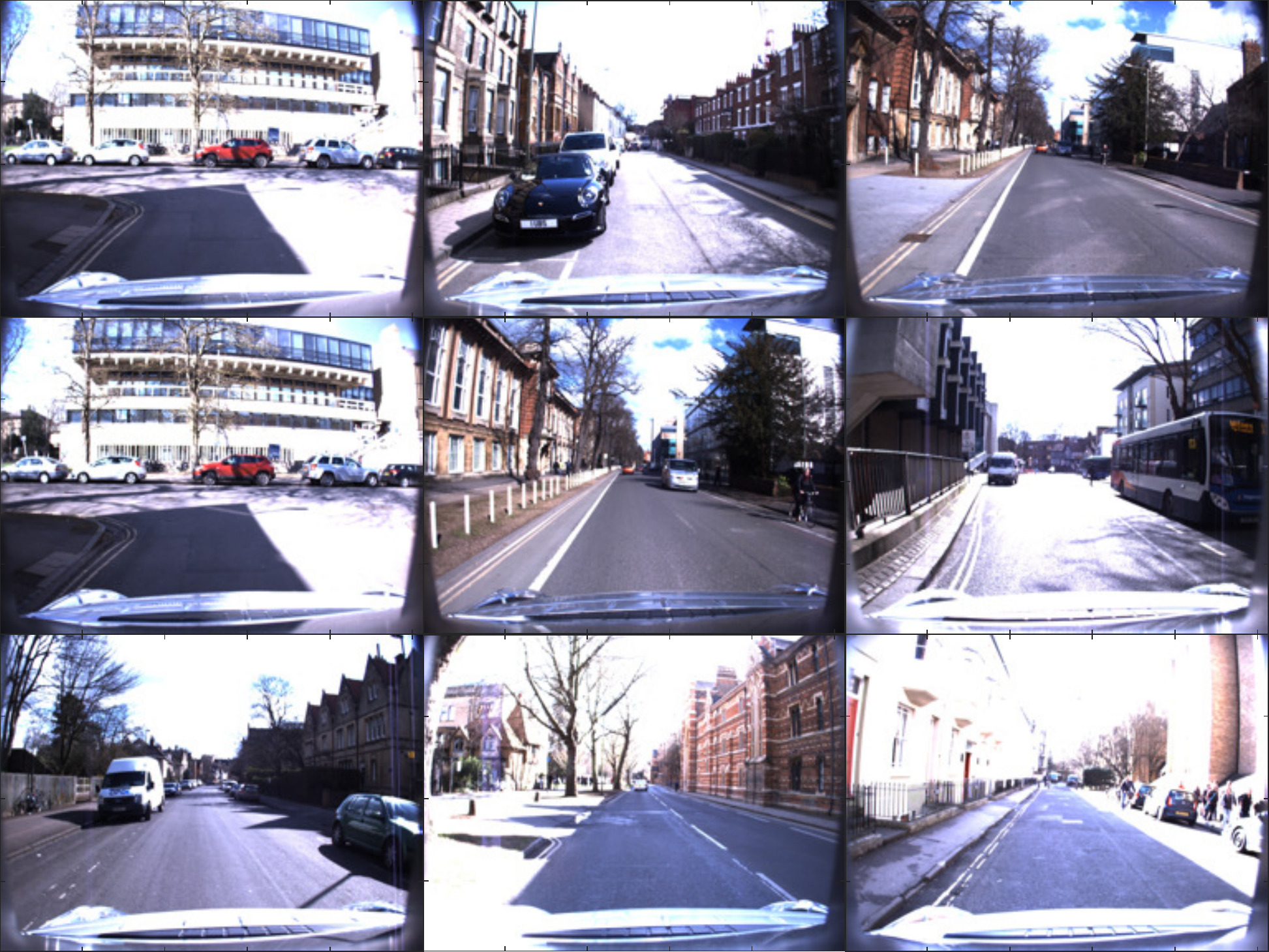}\\
	\end{tabular}
    \caption{\small Example images for source domain: overcast (top), and target domain: sunny (bottom).}
	\label{fig:sunny}
\end{figure}

The principal experiments were performed based on 9,000 training and 1,000 test images for each domain and 20 distinct location classes. We focus on the adaptation between overcast weather to sunny as displayed in Figure \ref{fig:sunny}.

The network architecture builds on AlexNet \cite{krizhevsky2012imagenet} and while we adapt the split between encoder and classifier / discriminator, the overall pipeline from image to location label is kept the same for all experiments, making this approach easy to apply to other common architectures such as FCN-VGG \cite{long2015fully}, which underlies the segmentation experiments. 

We determine the mean classification accuracy $P_T$ and standard deviation $\sigma$ 
for all tested hyperparameter configurations over 5 runs to investigate for GAN-typical training instability. All experiments are performed on an NVIDIA GTX TITAN GPU. All evaluations included in this work are run with the best found set of hyperparameters for the fixed parameters in each test.

The following subsections now address the questions posed in Section \ref{sec:introduction}.

\subsection{\textbf{Trade-offs}: How do we stabilise training and balance supervised and adversarial objectives?}
\label{sec:tradeoffs}

One of the most common issues with ADA is the potential inability to learn task-relevant and informative features as the domain confusion loss tries to reduce discriminator performance and can result in degenerate feature representations.

\subsubsection{Stabilising: Pretraining and Supervised Warm-up}

Initialising network architectures via pretraining of convolutional layers on large and diverse datasets is generally known to speed up the learning process as well as leading to better generalisation \cite{HuhAE16}.
Furthermore, in the context of ADA, it can be helpful to include a warmup phase where only the supervised task loss is used before switching on the adversarial loss.

We evaluate pretrained convolutional layers for AlexNet (based on ImageNet classification task~\cite{russakovsky2015imagenet}) as well as a supervised warmup phase of 15 epochs which, based on a small test run, has been found to work well in a large range of evaluation settings.

Table \ref{tab:pretrain} shows that for the given small number of samples, supervised warmup on its own does not suffice for a strong, classifier relevant feature representation therefore we build our following evaluation on pretrained convolutional layers as well as a 15 epochs supervised warmup phase.

\subsubsection{Balancing: Adversarial Loss}

\begin{figure}[ht]
	\centering
		\includegraphics[trim={1cm 0 0 0},width = 0.5\textwidth]{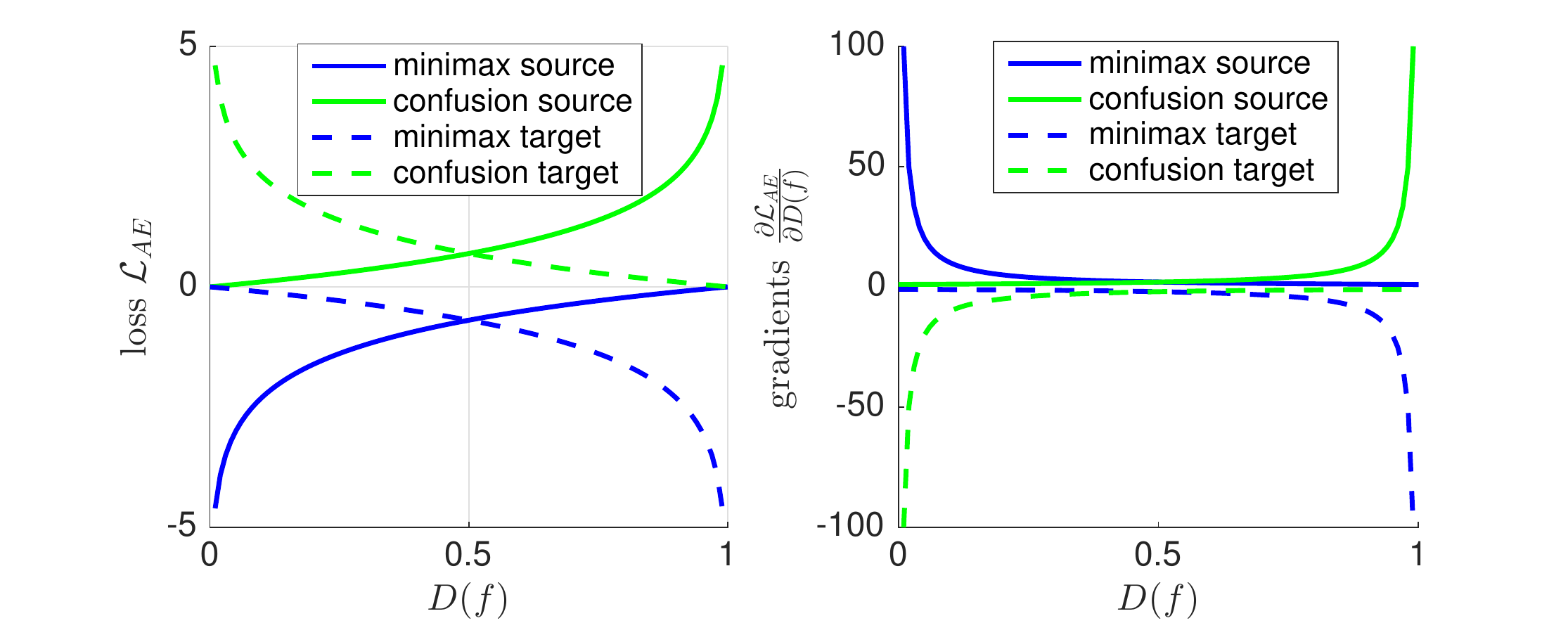}
    \caption{\small Effect of different encoder loss formulations on the gradients. The confusion loss results in significantly stronger gradients against a well performing discriminator (close to 1 for source data and 0 for target data). }
	\label{fig:encoder_loss}
\end{figure}

While the discriminator loss stays fixed across most recent work on ADA based on Equation \ref{eq:discr_loss}, the choice of the encoder loss varies and has been shown to have significant influence on convergence properties and stability of training in Generative Adversarial Networks~\cite{arjovsky2017towards}, which have strong similarities with ADA. The two main generator objectives to consider are the minimax formulation (negated discriminator loss) which is equal to the  gradient reversal layer \cite{Ganin2016} and the confusion loss~\cite{tzeng2015simultaneous}, which has found to prevent vanishing gradients with saturated discriminator but displays higher variance in the gradients \cite{arjovsky2017towards}. Additionally, the factor $\lambda$ is used to vary the relative strength of the adversarial loss in the overall training process to balance both objectives.

Figure \ref{fig:encoder_loss} shows minimax and confusion loss and gradients in dependence to the discriminator performance on source and target domain data. When the discriminator achieves high performance (close to 1 for source data and 0 for target data), the confusion loss results in significantly stronger gradients which will support a more stable adversarial training process. 

    \begin{table}[t]
        \small
        \vspace{3pt}
        \begin{tabular}{l|R{.15\textwidth}|R{.11\textwidth}|R{.11\textwidth}}
           \toprule
             & Supervised Warmup & Pretraining & Both \\
            \midrule
            $P_T [\%]$ & 52.68 & 72.03 & \textbf{82.03}	 \\
            $\sigma_T [\%]$ &6.57&	4.88& 2.09 \\
            \bottomrule
        \end{tabular}
        \caption{\small Evaluation of performance for network initialisation and warmup phase. The combination of both approaches leads to significant improvement}
        \label{tab:pretrain}
    \end{table}
    \begin{table}[!h]
        \small
        \begin{tabular}{l|R{.045\textwidth}|R{.045\textwidth}|R{.045\textwidth}|R{.045\textwidth}|R{.045\textwidth}|R{.045\textwidth}|R{.045\textwidth}}
           \toprule
              Loss& \multicolumn{7}{c}{Minimax Loss} \\
              $\lambda$ & $10^{-2}$ & $10^{-3}$ & $10^{-4}$  & $10^{-5}$ & $10^{-6}$ & $10^{-7}$ & $10^{-8}$ \\
            \midrule
            $P_T [\%]$ & 59.68&	55.22&	64.35&	73.28&	73.66&	76.32&	68.76\\
            $\sigma_T [\%]$ &	2.05&	3.40&	4.73&	3.76&	1.89&	2.13&	3.62\\
            \bottomrule    
            \toprule
            Loss& \multicolumn{7}{c}{Confusion Loss}  \\
            $\lambda$ & $10^{1}$ & $10^{0}$ & $10^{-1}$ & $10^{-2}$ & $10^{-3}$ & $10^{-4}$  & $10^{-5}$   \\
            \midrule
            $P_T [\%]$ & 76.04&	\textbf{82.03}&	71.89&	71.95&	71.28&	53.93&	37.77 \\
            $\sigma_T [\%]$ & 1.14&	2.09&	2.71&	2.00&	1.64&	29.06&	29.71\\
            \bottomrule
        \end{tabular}
        \caption{\small Evaluation of performance for minimax and confusion loss with variation of loss weighting. Optimal performance is achieved with the confusion.}
        \label{tab:loss}
    \end{table}
    \begin{table}[!h]
        \small
        \begin{tabular}{l|R{.045\textwidth}|R{.045\textwidth}|R{.045\textwidth}|R{.045\textwidth}|R{.045\textwidth}|R{.045\textwidth}|R{.045\textwidth}}
          \toprule
              Layer& 1 &  2 & 3 & 4  & 5 &  6 & 7 \\
            \midrule
            $P_T [\%] $ & 72.25&	74.14&	78.93&	\textbf{82.03}&	79.37&	79.72&	78.63\\
            $\sigma_T [\%]$ & 1.75&1.71&	2.90&	2.09&	1.07&	0.98&	0.84 \\
            \bottomrule
        \end{tabular}
        \caption{\small Evaluation of performance for the variation of the split layer between encoder and classifier/discriminator. The optimal split lies in the middle of the network. While a split to early in the architecture can lead to low performance, moving the split further towards the last layers only slowly affects performance.}
        \label{tab:splitlayer}
    \end{table}
    \begin{table}[!h]
        \scriptsize
        \begin{tabular}{l|R{.031\textwidth}|R{.031\textwidth}|R{.031\textwidth}|R{.031\textwidth}|R{.031\textwidth}|R{.031\textwidth}|R{.031\textwidth}|R{.031\textwidth}|R{.031\textwidth}}
            \toprule
            Layer & \multicolumn{3}{c|}{3} & \multicolumn{3}{c|}{4} & \multicolumn{3}{c}{5}\\
            Adapted \\ Capacity & -2 & 0 & 2 & -2  & 0 & 2 & -2 & 0 & 2 \\
            \midrule
            $P_T[\%]$ & 73.79 & 76.30 & 76.52  & 74.47 & \textbf{82.03} & 80.55 & 79.00 & 81.06 & 80.37 \\
            $\sigma_T[\%]$ & 6.08 & 3.94 & 3.17 & 2.62 & 2.09 & 1.94 & 0.67 & 0.53 & 1.66 \\
            \bottomrule
        \end{tabular}
        \caption{\small Evaluation of performance for adapting the discriminator capacity. For the optimal split layer, the best performance is achieved by not separately changing the discriminator architecture.}
        \label{tab:discriminator}
    \end{table}

The evaluation for the minimax loss in Table \ref{tab:loss} begins with considerably lower weights as higher values for $\lambda$ as $10^{-2}$ resulted in significant training instability risking the collapse of the target performance to chance (~5\%).
We have found the confusion loss significantly easier to apply as it renders the game between the adversarial modules more stable as displayed in Figure \ref{fig:encoder_loss}.
The original minimax-loss formulation was successfully employed in a number of recent works \cite{Ganin2016,bousmalis2016domain},
however, we have found the tuning process more complex and obtained significantly stronger results based on the confusion loss.

\subsection{\textbf{Architecture}: How can we directly adapt existing network architectures that are known to perform well?}
\label{sec:layer}

When applying ADA to a new task, one can greatly benefit from applying existing network architectures that have been optimised and proven to perform well on similar tasks.
This section addresses how to adapt existing architectures easily to incorporate the additional discriminator module.

The expressiveness and flexibility of encoder and discriminator directly influence the performance of our domain adaptation task.
While the flexibility of the discriminator limits the types of domain discrepancies that can be detected, the encoder  structure affects the efficiency of concealing the originating domain from the discriminator while generating relevant feature distributions for classification. 

\subsubsection{Choice of Split Layer}

The following evaluation focuses in particular on AlexNet \cite{krizhevsky2012imagenet},
which we adapt by providing the feature output of a particular layer additionally to a discriminator module.
Following Figure \ref{fig:ada}, 
the encoder module now includes all layers before the split layer and is shared for source and target domain.
We duplicate the architecture following the split layer for supervised module and discriminator and adapt the discriminator to output a single value per input as domain classifier.
By varying the \textit{split layer} with this approach 
we directly influence the capacity ratio between encoder and discriminator while keeping the overall number of layers for the supervised task fixed. 

The best possible split results in the middle of the network as both - encoder and discriminator - have enough capacity to fulfil their respective tasks (see Table \ref{tab:splitlayer}). An earlier split layer significantly reduces the encoders capacity to minimise mutual information with respect to the domain while encoding classification-relevant features. A later split decreases the expressiveness of the discriminator's density model such that less variations can be detected.

\subsubsection{Capacity of Discriminator}

While the choice of split layer as displayed above has strong influence on the effectiveness of domain adaptation, the approach might benefit from separately adapting the discriminator capacity to improve domain discrimination while keeping the rest of the architecture fixed.
In this context, we now evaluate the performance with respect to varying capacity of the discriminator by either adding or removing 2 fully connected layers before the final layer. 

Table \ref{tab:discriminator} presents that changing the capacity with the best split layer configuration (layer 4) leads to no improvement and it can be seen for all split layer configurations that trimming layers from the discriminator reduces target performance as the discriminator expressiveness is diminished. Furthermore, extending the discriminator leads to lowered accuracy as the model might overfit to domain discrepancy between both domain's training sets.

\subsection{\textbf{Performance}: Where does ADA work best and how do larger appearance changes influence performance?}
\label{sec:applications}

To provide bounds for the performance of the approach we evaluate the classification accuracy of the classifier only trained on training data from the source domain as lower bound and only with training data from the target domain as upper bound. 
Note, all models are initialised with pretrained convolutional layers pretrained on ImageNet~\cite{russakovsky2015imagenet}.

\begin{figure}[t]
	\centering
	\begin{tabular}{c}
		\includegraphics[width = 0.43\textwidth]{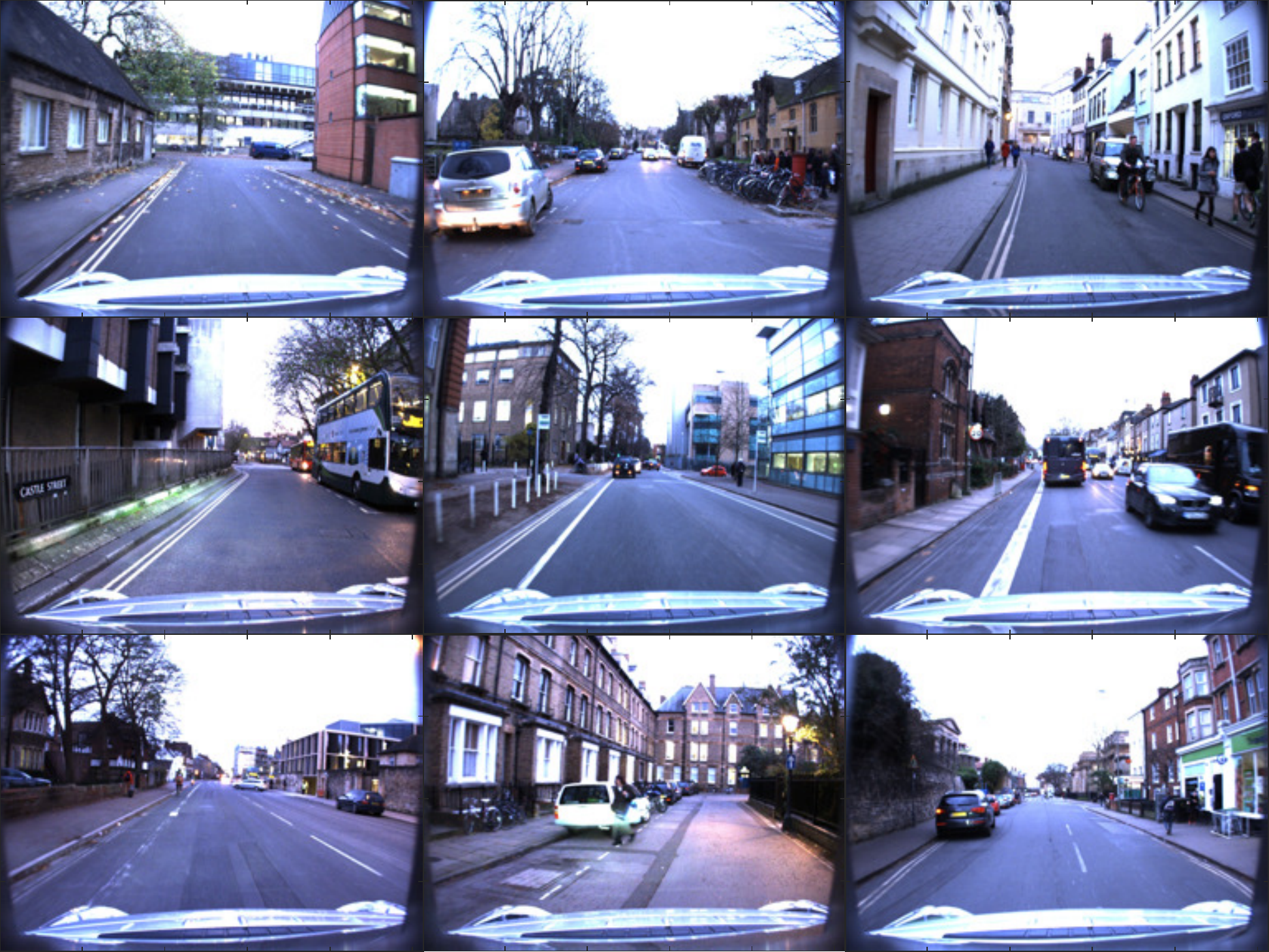}\\
        \includegraphics[width = 0.43\textwidth]{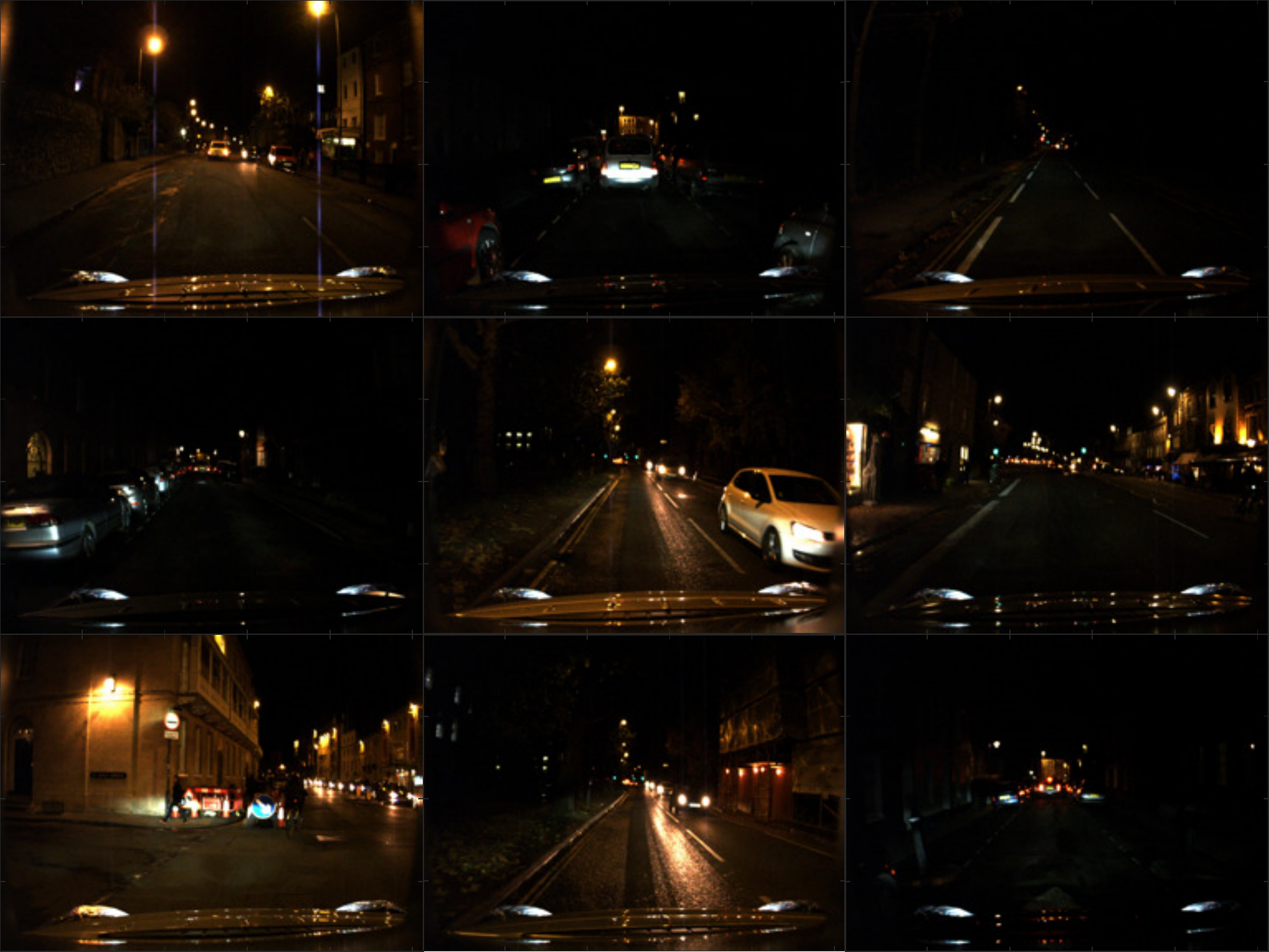}\\
	\end{tabular}
    \caption{\small Example images for source domain: day (top), and target domain: night (bottom).}
	\label{fig:night}
\end{figure}

As the complexity for unsupervised domain adaptation directly correlates with the difference between the distributions, we evaluate benefits of the approach on the following source-target pairs \textit{Sunny - Overcast} and \textit{Day - Night} with the latter representing a significantly more complex transfer as visualised in Figure \ref{fig:night}.

\begin{table}[!h]
    \centering
    \small
    \begin{tabular}{l|c|c|c|c}
        \multicolumn{5}{c}{Classification} \\
        \toprule
        & domains & \parbox[t][][t]{45pt}{\centering AlexNet}  & \parbox[t][][t]{45pt}{\centering AlexNet w ADA}& \parbox[t][][t]{45pt}{\centering AlexNet w target labels \footnotemark[1]} \\
        \midrule
        $P_T[\%]$ & overcast-sunny & 67.95$\mypm$1.02 & \textbf{82.03}$\mypm$2.09 & (87.96$\mypm$1.40)  \\
        $P_T[\%]$ & day-night &	26.79$\mypm$4.90  & \textbf{30.21}$\mypm$4.94 &	(90.42$\mypm$1.04) \\
        \bottomrule
    \end{tabular}
    \begin{tabular}{l|c|c|c|c}
        \multicolumn{5}{c}{Free-Space Segmentation} \\
        \toprule
          & \parbox[t][][t]{48pt}{\centering FCV-VGG16}  & \parbox[t][][t]{48pt}{\centering FCN-VGG16 \\ w patch-ADA}& \parbox[t][][t]{48pt}{\centering FCN-VGG16 \\ w ADA}  & \parbox[t][][t]{48pt}{\centering FCN-VGG16 \\ w target labels \footnotemark[1]}  \\
        \midrule
        $P_T[\%]$ & 75.12$\mypm$0.76&	67.68$\mypm$4.25 &\textbf{85.27}$\mypm$1.03&	(93.94$\mypm$0.84)	  \\
        \bottomrule
    \end{tabular}
    \vspace{3pt}
    \caption{\small Evaluation of the performance gains based on ADA for the surrogate classification task and free-space segmentation in the context of autonomous driving. ADA leads to significantly higher accuracy in all transfer domains and approaches in scenarios with minor shift towards the performance with known target labels. In the day-night scenario based on a strong difference in appearance, the approach only leads to minor benefits. The segmentation task measures performance $P_T$ as mean average precision.}
    \label{tab:main tasks}
\end{table}
\footnotetext[1]{Performance with available training labels on the target domain. This serves as upper bound for performance.}

By increasing the domain shift occurring between source an target data, we can investigate limitations of the ADA. While the approach leads to performance gains in both scenarios, it is clearly better suited to address domains with limited shift, such as the overcast-sunny scenario, and performs significantly closer to the upper bound obtained from training on labelled target data as displayed in Table \ref{tab:main tasks}.

\subsubsection{Improving Performance in Source Domain} 
To fully evaluate the suitability of ADA for long-term robotics applications, this section aims at investigating the source performance $P_S$ as the original labelled domain is still of relevance.
In addition to improving target performance, ADA can acts as a regulariser to improve generalisation and test performance in the source domain as represented in Table~\ref{tab:sourceperform}.
This gain in performance is only possible as long as both domains have significant structural similarities and the dominant variations are shared by both domains. With increasing discrepancy between the domains, it can however reduce performance on the source domain as it might diminish information that helps it to generalise in source but not in the target domain, as can be seen with respect to the day-night adaptation in Table \ref{tab:sourceperform}.

\begin{table}[!h]
    \centering
    \small
    \begin{tabular}{l|c|c|c}
       \toprule
            & domains & trained on source &  ADA\\
        \midrule
        $P_S [\%]$ & overcast-sunny & 87.83$\mypm$1.56 & \textbf{90.82}$\mypm$1.45 \\
        $P_S [\%]$ & day-night & \textbf{90.04}$\mypm$3.01 & 89.16$\mypm$1.78 \\
        \bottomrule
    \end{tabular}
    \vspace{3pt}
    \caption{\small Evaluation of the regularising effect of ADA on source performance. For minor appearance difference between the domains, the approach can improve source performance by acting as a regulariser. However, under stronger shifts such as the day-night transfer it can lead to reduced source performance.}
    \label{tab:sourceperform}
\end{table}

\vspace{1mm}
\section{Segmentation Task}
\label{sec:segmentation}

Following the optimisation for our surrogate task, we now apply the distilled insights for optimising ADA to the task of free space segmentation as possible input data for motion planning systems.
We use the fully convolutional FCN-VGG16~\cite{long2015fully} architecture, which is split - similar to the classification tests - into encoder and classifier/discriminator. We set the split layer towards the middle of the architecture after the 4th maxpool operation (see \cite{long2015fully} for the exact architecture) with fixed capacity of the discriminator architecture and apply the confusion loss. 

Both source and target datasets include 1000 training and 100 test images based on a midday to early evening adaptation scenario such that the domain shift is intuitively smaller than in the full day to night transfer from section \ref{sec:applications}. The segmentation labels are generated for free-space/obstacles following the approach of Barnes et al~\cite{BarnesMP16}.

\begin{figure}[h]
	\centering
	\begin{tabular}{c c}
		\includegraphics[width = 0.23\textwidth]{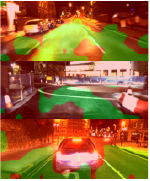}&
        \includegraphics[width = 0.23\textwidth]{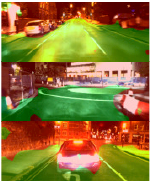}\\
        FCV-VGG16 & FCV-VGG16 w ADA \\
        \includegraphics[width = 0.23\textwidth]{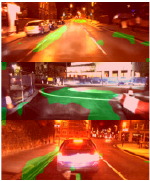}&
        \includegraphics[width = 0.23\textwidth]{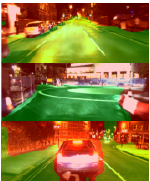}\\
        FCV-VGG16 w patch-ADA & FCV-VGG16 w target labels\footnotemark[1] \\
	\end{tabular}
    \caption{\small Examples of segmentation results in the target domain. Green and red represent free-space and obstacles respectively. ADA leads to significant performance gains for the unsupervised target domain while patch-ADA fails to increase accuracy. The scenario with available target labels serves as upper bound for the performance when supervised information would be available.}
	\label{fig:segmentation}
\end{figure}

The domain adapted segmentation network performs significantly better than the basic supervised approach and is able to bridge the gap towards performance with available labels in the target domain as displayed in Figure \ref{fig:segmentation}.

As the segmentation output of the approach has a limited receptive field for each pixel location we additionally evaluate a patch-wise discriminator in line with research on image-to-image translation with conditional Generative Adversarial Networks (cGAN)~\cite{isola2016image}.
This approach enables to keep the size of the receptive field fixed between the supervised and adversarial task.
While Isola et al~\cite{isola2016image} have found the patch discriminator to work better for their task based on cGANs, it resulted in reduced accuracy for our application on ADA.

\section{Limitations of Unsupervised Adversarial Domain Adaptation}
\label{sec:limits}

While we have shown domain adaptation to be beneficial in many scenarios, it must be marked that the current approach finds its limitations when the differences between source and target domains are too severe.
As exemplified by the day to night transfer scenario, the approach still leads to some improvement but with stronger variation in the underlying domains, the adversarial encoder loss might even lead to reduced performance if weighted improperly against the supervised loss. If the adversarial loss dominates in such situations the encoder features might lose more information relevant for the supervised task.

Finally, to overcome the limitation of significant domain shifts, semi-supervised approaches can be employed to incorporate further structure and align the conditional distributions over feature representations given labels comparable to \cite{tzeng2015simultaneous}.

\vspace{1mm}
\section{Discussion}
\label{sec:discussion}

Adversarial training frameworks such as ADA tend to be notoriously hard to train. However, a limited number hyperparameters has strong influence for a given problem and can be adapted to stabilise and optimise the training process.

While the detailed performance depends on architecture and task, we found the principal factors for optimising performance to be:
\begin{itemize}
    \item Using relevant initialisation including supervised warmup helps to guide the training process.
    \item Applying the confusion loss for the encoder enables better balancing and stabilising.
    \item The optimum position for the split layer is mid network. Particularly the application in earlier layers can significantly reduce the benefits.
\end{itemize}

Additional to the main evaluation above, we tested the influence of other advancements from the related GAN framework. It was found that neither mini-batch discriminator \cite{salimans2016improved} nor discriminator noise \cite{salimans2016improved} brought significant advances. 
This is justified as the former addresses generator mode collapse which in the context of ADA will be less of a problem as the supervised loss guides towards more versatile solutions. The latter seems to have negligible influence in comparison to the GAN framework as the discriminator's task is of higher complexity and with reasonable loss settings, the risk of the discriminator saturating is minimal.

As a side note, we find it helpful in this adversarial training framework to apply gradient clipping to prevent abrupt instabilities when one of the adversarial models finds a strong exploit.

\vspace{2mm}
\section{Conclusion and Future Work}
\label{sec:conclusion}

In this paper, we cast the common challenge of appearance change in outdoor robotics as an unsupervised domain adaptation problem and extend recent adversarial paradigms for the adaptation of popular, existing architectures leading to performance benefits in unlabelled target domains.
While instabilities of adversarial training can inhibit the extension to large scale problems, our extensive tests on a surrogate task with moderate complexity expose the most significant factors of influence and enable application on the full-scale path segmentation task for autonomous driving. 
With this straightforward framework we hope to pave the way for further application on real world tasks, in particular in the context of autonomous mobility where strong structural similarities of feature distribution can exist between different source and target domains, e.g. based on spatial overlay of driven routes and terrains.

Beyond dealing with appearance change we see many potential applications of ADA in robotics where sensor modalities may change, or even transferring models from a simulated virtual environment to improve their performance in the real world.

Beyond dealing with appearance change we see many potential applications of ADA in robotics, whether it be changes in sensor modalities, or transferring models from a simulated virtual environment to improve their performance in the real world.

Lastly, as the alignment of marginal distributions was observed to be significantly more successful between similar domains, there is a strong opportunity to explore curriculum based variations of ADA with a focus on gradual alignment in continually changing environments. The approach is of relevance in particular for scenarios with continuous application of mobile platforms as well as in lifelong learning tasks.

\section*{Acknowledgment}
The authors would like to acknowledge the support of the UK’s Engineering and Physical Sciences Research Council (EPSRC) through the Programme Grant DFR01420 and the Doctoral Training Award (DTA) as well as the support of the Hans-Lenze-Foundation.

\bibliographystyle{unsrt}
\bibliography{main}

\begin{thebibliography}{10}

\bibitem{paton2015s}
Michael Paton, Kirk MacTavish, Chris~J Ostafew, and Timothy~D Barfoot.
\newblock It's not easy seeing green: Lighting-resistant stereo visual teach \&
  repeat using color-constant images.
\newblock In {\em Robotics and Automation (ICRA), 2015 IEEE International
  Conference on}, pages 1519--1526. IEEE, 2015.

\bibitem{ChurchillIJRR2013}
Winston Churchill and Paul Newman.
\newblock {E}xperience-based {N}avigation for {L}ong-term {L}ocalisation.
\newblock {\em The International Journal of Robotics Research (IJRR)}, 2013.

\bibitem{maddern2014illumination}
Will Maddern, Alex Stewart, Colin McManus, Ben Upcroft, Winston Churchill, and
  Paul Newman.
\newblock Illumination invariant imaging: Applications in robust vision-based
  localisation, mapping and classification for autonomous vehicles.
\newblock In {\em Proceedings of the Visual Place Recognition in Changing
  Environments Workshop, IEEE International Conference on Robotics and
  Automation (ICRA), Hong Kong, China}, volume~2, page~3, 2014.

\bibitem{lai20093d}
Kevin Lai and Dieter Fox.
\newblock 3d laser scan classification using web data and domain adaptation.
\newblock In {\em Robotics: Science and Systems}, volume~2, 2009.

\bibitem{tzeng2015towards}
Eric Tzeng, Coline Devin, Judy Hoffman, Chelsea Finn, Xingchao Peng, Sergey
  Levine, Kate Saenko, and Trevor Darrell.
\newblock Towards adapting deep visuomotor representations from simulated to
  real environments.
\newblock {\em arXiv preprint arXiv:1511.07111}, 2015.

\bibitem{daume2006domain}
Hal Daume~III and Daniel Marcu.
\newblock Domain adaptation for statistical classifiers.
\newblock {\em Journal of Artificial Intelligence Research}, 26:101--126, 2006.

\bibitem{sugiyama2008direct}
Masashi Sugiyama, Shinichi Nakajima, Hisashi Kashima, Paul~V Buenau, and
  Motoaki Kawanabe.
\newblock Direct importance estimation with model selection and its application
  to covariate shift adaptation.
\newblock In {\em Advances in neural information processing systems}, pages
  1433--1440, 2008.

\bibitem{goodfellow2014generative}
Ian Goodfellow, Jean Pouget-Abadie, Mehdi Mirza, Bing Xu, David Warde-Farley,
  Sherjil Ozair, Aaron Courville, and Yoshua Bengio.
\newblock Generative adversarial nets.
\newblock In {\em Advances in neural information processing systems}, pages
  2672--2680, 2014.

\bibitem{ShrivastavaPTSW16}
Ashish Shrivastava, Tomas Pfister, Oncel Tuzel, Josh Susskind, Wenda Wang, and
  Russ Webb.
\newblock Learning from simulated and unsupervised images through adversarial
  training.
\newblock {\em CoRR}, abs/1612.07828, 2016.

\bibitem{bousmalis2016domain}
Konstantinos Bousmalis, George Trigeorgis, Nathan Silberman, Dilip Krishnan,
  and Dumitru Erhan.
\newblock Domain separation networks.
\newblock In {\em Advances in Neural Information Processing Systems}, pages
  343--351, 2016.

\bibitem{Ganin2016}
Yaroslav Ganin, Evgeniya Ustinova, Hana Ajakan, Pascal Germain, Hugo
  Larochelle, Fran{\c{c}}ois Laviolette, Mario Marchand, Victor Lempitsky, Urun
  Dogan, Marius Kloft, Francesco Orabona, and Tatiana Tommasi.
\newblock {Domain-Adversarial Training of Neural Networks}.
\newblock {\em Journal of Machine Learning Research}, 17:1--35, 2016.

\bibitem{krizhevsky2012imagenet}
Alex Krizhevsky, Ilya Sutskever, and Geoffrey~E Hinton.
\newblock Imagenet classification with deep convolutional neural networks.
\newblock In {\em Advances in neural information processing systems}, pages
  1097--1105, 2012.

\bibitem{long2015fully}
Jonathan Long, Evan Shelhamer, and Trevor Darrell.
\newblock Fully convolutional networks for semantic segmentation.
\newblock In {\em Proceedings of the IEEE Conference on Computer Vision and
  Pattern Recognition}, pages 3431--3440, 2015.

\bibitem{Goodfellow17}
Ian~J. Goodfellow.
\newblock {NIPS} 2016 tutorial: Generative adversarial networks.
\newblock {\em CoRR}, abs/1701.00160, 2017.

\bibitem{arjovsky2017towards}
Martin Arjovsky and L{\'e}on Bottou.
\newblock Towards principled methods for training generative adversarial
  networks.
\newblock In {\em NIPS 2016 Workshop on Adversarial Training. In review for
  ICLR}, volume 2016, 2017.

\bibitem{ratnasingam2010study}
Sivalogeswaran Ratnasingam and Steve Collins.
\newblock Study of the photodetector characteristics of a camera for color
  constancy in natural scenes.
\newblock {\em JOSA A}, 27(2):286--294, 2010.

\bibitem{corke2013dealing}
Peter Corke, Rohan Paul, Winston Churchill, and Paul Newman.
\newblock Dealing with shadows: Capturing intrinsic scene appearance for
  image-based outdoor localisation.
\newblock In {\em Intelligent Robots and Systems}, pages 2085--2092. IEEE,
  2013.

\bibitem{milford2012seqslam}
Michael~J Milford and Gordon~F Wyeth.
\newblock Seqslam: Visual route-based navigation for sunny summer days and
  stormy winter nights.
\newblock In {\em Robotics and Automation (ICRA), 2012 IEEE International
  Conference on}, pages 1643--1649. IEEE, 2012.

\bibitem{neubert2013appearance}
Peer Neubert, Niko Sunderhauf, and Peter Protzel.
\newblock Appearance change prediction for long-term navigation across seasons.
\newblock In {\em Mobile Robots (ECMR), 2013 European Conference on}, pages
  198--203. IEEE, 2013.

\bibitem{ben2007analysis}
Shai Ben-David, John Blitzer, Koby Crammer, Fernando Pereira, et~al.
\newblock Analysis of representations for domain adaptation.
\newblock {\em Advances in neural information processing systems}, 19:137,
  2007.

\bibitem{bousmalis2016unsupervised}
Konstantinos Bousmalis, Nathan Silberman, David Dohan, Dumitru Erhan, and Dilip
  Krishnan.
\newblock Unsupervised pixel-level domain adaptation with generative
  adversarial networks.
\newblock {\em arXiv preprint arXiv:1612.05424}, 2016.

\bibitem{Hong2017}
Sungeun Hong, Woobin Im, Jongbin Ryu, and Hyun~S Yang.
\newblock Sspp-dan: Deep domain adaptation network for face recognition with
  single sample per person.
\newblock {\em arXiv preprint arXiv:1702.04069}, 2017.

\bibitem{Long0J16a}
Mingsheng Long, Jianmin Wang, and Michael~I. Jordan.
\newblock Deep transfer learning with joint adaptation networks.
\newblock {\em CoRR}, abs/1605.06636, 2016.

\bibitem{SunFS16}
Baochen Sun, Jiashi Feng, and Kate Saenko.
\newblock Correlation alignment for unsupervised domain adaptation.
\newblock {\em CoRR}, abs/1612.01939, 2016.

\bibitem{ajakan2014domain}
Hana Ajakan, Pascal Germain, Hugo Larochelle, Fran{\c{c}}ois Laviolette, and
  Mario Marchand.
\newblock Domain-adversarial neural networks.
\newblock {\em arXiv preprint arXiv:1412.4446}, 2014.

\bibitem{tzeng2015simultaneous}
Eric Tzeng, Judy Hoffman, Trevor Darrell, and Kate Saenko.
\newblock Simultaneous deep transfer across domains and tasks.
\newblock In {\em Proceedings of the IEEE International Conference on Computer
  Vision}, pages 4068--4076, 2015.

\bibitem{KamnitsasBLNSKM16}
Konstantinos Kamnitsas, Christian~F. Baumgartner, Christian Ledig, Virginia
  F.~J. Newcombe, Joanna~P. Simpson, Andrew~D. Kane, David~K. Menon, Aditya
  Nori, Antonio Criminisi, Daniel Rueckert, and Ben Glocker.
\newblock Unsupervised domain adaptation in brain lesion segmentation with
  adversarial networks.
\newblock {\em CoRR}, abs/1612.08894, 2016.

\bibitem{BarnesMP16}
Dan Barnes, William~P. Maddern, and Ingmar Posner.
\newblock Find your own way: Weakly-supervised segmentation of path proposals
  for urban autonomy.
\newblock {\em CoRR}, abs/1610.01238, 2016.

\bibitem{RobotCarDatasetIJRR}
Will Maddern, Geoff Pascoe, Chris Linegar, and Paul Newman.
\newblock {1 Year, 1000km: The Oxford RobotCar Dataset}.
\newblock {\em The International Journal of Robotics Research (IJRR)}, 2016.

\bibitem{HuhAE16}
Mi{-}Young Huh, Pulkit Agrawal, and Alexei~A. Efros.
\newblock What makes imagenet good for transfer learning?
\newblock {\em CoRR}, abs/1608.08614, 2016.

\bibitem{russakovsky2015imagenet}
Olga Russakovsky, Jia Deng, Hao Su, Jonathan Krause, Sanjeev Satheesh, Sean Ma,
  Zhiheng Huang, Andrej Karpathy, Aditya Khosla, Michael Bernstein, et~al.
\newblock Imagenet large scale visual recognition challenge.
\newblock {\em International Journal of Computer Vision}, 115(3):211--252,
  2015.

\bibitem{isola2016image}
Phillip Isola, Jun-Yan Zhu, Tinghui Zhou, and Alexei~A Efros.
\newblock Image-to-image translation with conditional adversarial networks.
\newblock {\em arXiv preprint arXiv:1611.07004}, 2016.

\bibitem{salimans2016improved}
Tim Salimans, Ian Goodfellow, Wojciech Zaremba, Vicki Cheung, Alec Radford, and
  Xi~Chen.
\newblock Improved techniques for training gans.
\newblock In {\em Advances in Neural Information Processing Systems}, pages
  2226--2234, 2016.

\end{thebibliography}

\end{document}